\documentclass{article}

\usepackage[utf8]{inputenc} 
\usepackage[T1]{fontenc}    
\usepackage{hyperref}       
\usepackage{url}            
\usepackage{booktabs}       
\usepackage{amsfonts}       
\usepackage{nicefrac}       
\usepackage{microtype}      
\usepackage{authblk}

\usepackage{xcolor}
\usepackage{url}
\usepackage{latexsym}

\usepackage{mathtools}
\usepackage{calc}
\usepackage{amsmath, amsthm, amssymb, amsfonts}
\PassOptionsToPackage{numbers, compress}{natbib}
\usepackage[final]{nips_2017}

\usepackage{graphicx}
\def\shrug{\texttt{\raisebox{0.75em}{\char`\_}\char`\\\char`\_\kern-0.5ex(\kern-0.25ex\raisebox{0.25ex}{\rotatebox{45}{\raisebox{-.75ex}"\kern-1.5ex\rotatebox{-90})}}\kern-0.5ex)\kern-0.5ex\char`\_/\raisebox{0.75em}{\char`\_}}}

\title{Attending to All Mention Pairs for \break Full Abstract Biological Relation Extraction}

\author[1]{\bf Patrick Verga}
\author[1]{\qquad \bf Emma Strubell}
\author[2]{\qquad \bf Ofer Shai}
\author[1]{\qquad \bf Andrew McCallum}
\affil[1]{College of Information and Computer Sciences \protect\\
University of Massachusetts Amherst}
\affil[2]{Chan Zuckerberg Initiative \protect\\
\texttt{\{pat, strubell, mccallum\}@cs.umass.edu, oshai@chanzuckerberg.com}}

\date{}

\begin{document}

\maketitle
\begin{abstract}
Most work in relation extraction forms a prediction by looking at a short span of text within a single sentence containing a single entity pair mention.  However, many relation types, particularly in biomedical text, are expressed across sentences or require a large context to disambiguate. We propose a model to consider all mention and entity pairs simultaneously in order to make a prediction. We encode full paper abstracts using an efficient self-attention encoder and form pairwise predictions between all mentions with a bi-affine operation. An entity-pair wise pooling aggregates mention pair scores to make a final prediction while alleviating training noise by performing within document multi-instance learning. We improve our model's performance by jointly training the model to predict named entities and adding an additional corpus of weakly labeled data. We demonstrate our model's effectiveness by achieving the state of the art on the Biocreative V Chemical Disease Relation dataset for models without KB resources, outperforming ensembles of models which use hand-crafted features and additional linguistic resources.

\end{abstract}

\section{Introduction}
\label{sec:intro}
With few exceptions, nearly all work in relation extraction focuses on classifying a short span of text within a single sentence containing a single entity pair mention. However, relationships between entities are often expressed across sentence boundaries or require a larger context to disambiguate. For example, in the Biocreative V CDR dataset (Section \ref{sec:results}), 30\% of relations are expressed across sentence boundaries, such as in the following excerpt:

\begin{quote}
\small
\textit{Treatment of psoriasis with \textbf{\textcolor{blue}{azathioprine}}}. \textbf{\textcolor{blue}{Azathioprine}} treatment benefited 19 (66\%) out of 29 patients suffering from severe psoriasis. Haematological complications were not troublesome and results of biochemical liver function tests remained normal. Minimal cholestasis was seen in two cases and portal \textbf{\textcolor{red}{fibrosis}} of a reversible degree in eight. Liver biopsies should be undertaken at regular intervals if \textbf{\textcolor{blue}{azathioprine}} therapy is continued so that structural liver damage may be detected at an early and reversible stage.
\end{quote}
Though the entities' mentions never occur in the same sentence, the above example expresses that the chemical entity \emph{azathioprine} can cause the side effect \emph{fibrosis}. In order to extract relations from such text, between entities with relations expressed across sentence boundaries, we propose Bi-affine Relation Attention Networks (BRAN) which predict relationships between all mention pairs within a document simultaneously. We efficiently encode full-paper abstracts using self-attention over byte-pair encoded sub-word tokens. This allows the model to consider a wider context between distant mention pairs, as well as correlations between multiple mentions. Making simultaneous predictions also allows us to apply within-document multi-instance learning, leveraging document level annotation by alleviating noise caused by a lack of mention level annotation. We demonstrate state of the art performance for a model using no external knowledge base resources in experiments on the Biocreative V CDR dataset. 

\section{Model}
\label{sec:model}

Our model first contextually encodes input token embeddings. These contextual embeddings are used to predict both entities and relations. The relation extraction module converts each token to a \emph{head} and \emph{tail} representation. These representations are used to form mention-pair predictions using a bi-affine operation with respect to learned relation embeddings. Finally the mention-level predictions are pooled to form an entity-level prediction.

\begin{figure}
  \begin{minipage}[c]{0.45\textwidth}
    \includegraphics[scale=.65]{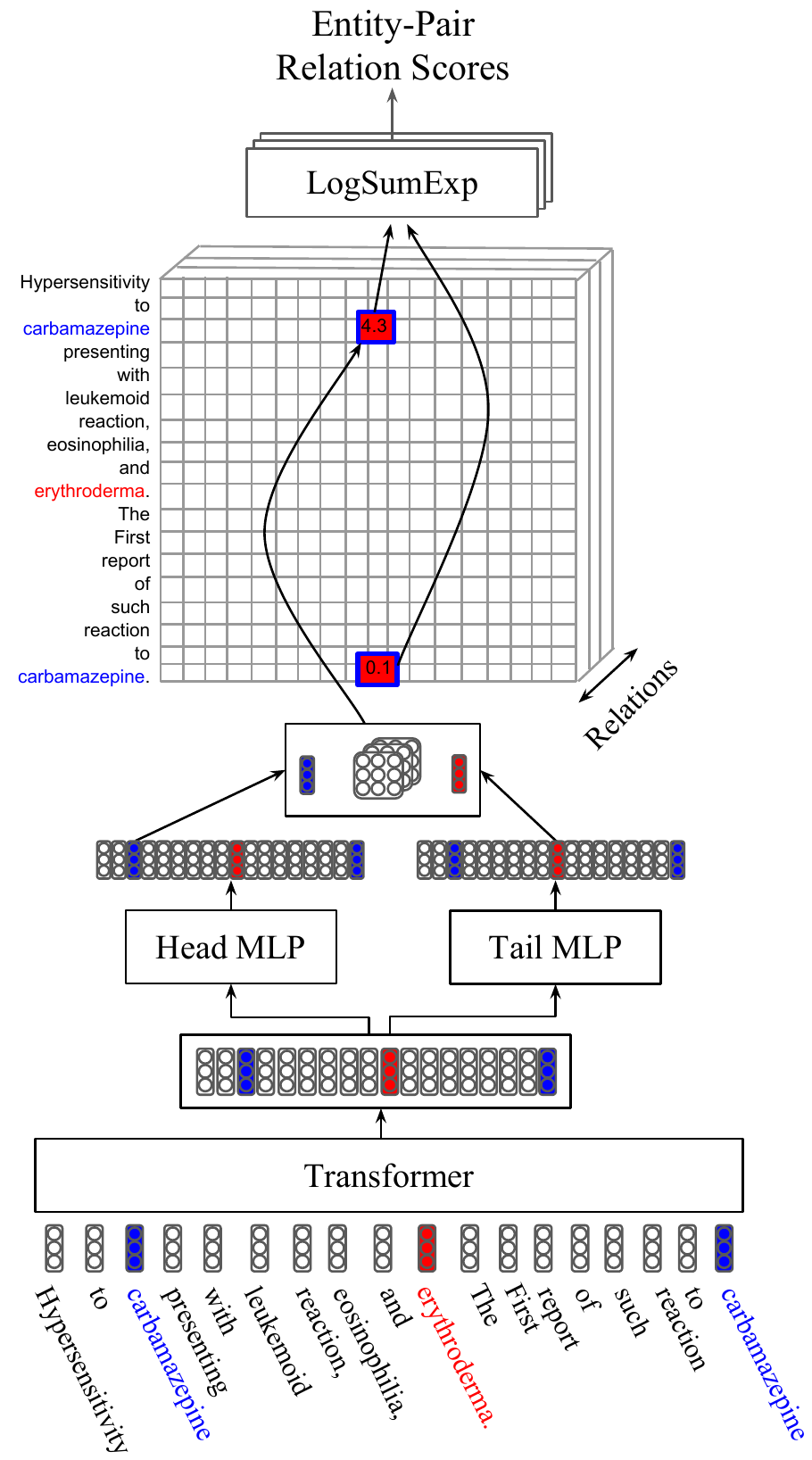}
  \end{minipage}\hfill
  \begin{minipage}[c]{0.45\textwidth}
    \caption{The relation extraction architecture. Inputs are contextually encoded using the Transformer\citep{vaswani2017attention}. Each transformed token is then passed through a \emph{head} and \emph{tail} MLP to produce two separate versions of each token. A bi-affine operation is then performed between each \emph{head} and \emph{tail} token with respect to each relation's embedding matrix, producing a pairwise-relation affinity tensor. Finally, the scores for cells corresponding to the same entity pair are pooled with a separate LogSumExp operation for each relation to get a final score. The colored tokens illustrate calculating the score for a given pair of entities; The model is only given entity information when gathering scores to pool from the affinity matrix. \label{fig:transformer_all_pairs}}
  \end{minipage}
\end{figure}

\subsection {Inputs}
Our model takes in a sequence of $n$ token embeddings in $\mathbb{R}^d$. Because the original Transformer has no recurrence, convolutions, or other mechanism of modeling position information, the model relies on positional embeddings which are added to the input token embeddings\footnote{Even though our final model incorporates some convolutions, we retain the position embeddings}. We learn position embedding matrix $P^{m\times d}$ which contains a separate $d$ dimensional embedding for each position, limited to $m$ possible positions. Our final input representation for token $x_i$ is:
\begin{align*}
x_i=s_i+p_i
\end{align*}

Where $s_i$ is the token embedding for $x_i$ and $p_i$ is the positional embedding for the $i$th position. If $i$ exceeds $m$, we use a randomly initialized vector in place of $p_i$.

We tokenize the text using byte pair encoding \citep{gage1994new,sennrich2015neural} which is well suited for biological data for a number of reasons. First, biological entities often have unique mentions made up of meaningful subcomponents, such as `1,2-dimethylhydrazine'. By learning sub-word representations the model is able to make predictions on rare or unknown words. Additionally, tokenization of chemical entities is challenging, lacking a universally agreed upon algorithm \citep{krallinger2015chemdner}. 

Byte pair encoding constructs a vocabulary of sub-word pieces beginning with single characters. Then, the algorithm iteratively merges the most frequent co-occurring tokens into a new token, which is added to the vocabulary. This procedure continues until a pre-defined vocabulary size is met. 

\subsection {Transformer}\label{sec:model_multihead_attention}
We use the Transformer self-attention model \citep{vaswani2017attention} to encode tokens by aggregating over their context in the entire sequence. The Transformer is made up of $B$ blocks. Each Transformer block, $\mathrm{Transformer}_k$, has its own set of parameters and is made up of two subcomponents, multi-head attention and a series of convolutions. The output $b_i^{(k)}$ of block $k$ is connected to its input $b_i^{(k-1)}$ with a residual connection \citep{he2016deep}:

\begin{align*}
b_i^{(k)}=b_i^{(k-1)}+\mathrm{Transformer}_k(b_i^{(k-1)})
\end{align*}

\subsubsection{Multi-headed Attention\label{sec:multihead-attention}}
Multi-head attention applies self-attention multiple times over the same inputs using separate parameters (attention heads) and combines the results, as an alternative to  applying one pass of attention with more parameters. The intuition behind this modeling decision is that dividing the attention into multiple heads make it easier for the model to learn to attend to different types of relevant information with each head. The self-attention updates input $b_i^{(k)}$ by aggregating information for all tokens in the sequence weighted by their importance. 

Each input is projected to a key, value, and query, using separate affine transformations with $\mathrm{ReLU}$ activations. Where $k$, $v$, and $q$, each in $\mathbb{R}^{\frac{d}{h}}$ where $h$ is the number of heads. The attention weights $a_{ijh}$ are computed as scaled dot-product attention as:


\begin{align*}
a_{ijh} &= \sigma\left(\frac{q_{ih}^T k_{jh}}{\sqrt{d}}\right) \\ 
o_{ih} &= \sum_j v_{jh} \odot a_{ijh} \\
\end{align*}

Where $\odot$ is element-wise multiplication and $\sigma$ indicates a softmax along the $jth$ dimension. The scaled attention is meant to aid optimization by flattening the softmax and better distributing the gradients \citep{vaswani2017attention}.

The outputs of the individual heads of the multi-headed attention are concatenated, denoted $[;]$, into $o_i$. All layers in the network use residual connections between the output of the multi-headed attention and its input. Layer normalization \citep{ba2016layer}, denoted LN$(\cdot)$, is then applied to the output. 
 \begin{align*}
o_i &= [o_1;...;o_h] \\
m_i &= \mathrm{LN}(b_i^{(k)} + o_i)
\end{align*}

\subsubsection{Feed-Forward\label{sec:model_feedforward}}

The second part of the transformer block is a stack of convolutional layers. The sub-network used in \citet{vaswani2017attention} uses two width-1 convolutions. We add a third middle layer with kernel width 5, which we found to perform better. Many relations are expressed concisely by the immediate local context (``Michele's husband Barack'', ``labetalol -induced hypotension''). Adding in this explicit n-gram modeling is meant to ease the burden on the model to learn to attend to the local features entirely on its own. We use $C_w$ to denote a convolutional layer with convolutional kernel width $w$. Then the convolutional portion of the transformer block is given by:

\begin{align*}
t_i^{(0)} &= \mathrm{ReLU}(C_1(m_i)) \\
t_i^{(1)} &= \mathrm{ReLU}(C_5(t_i^{(0)})) \\
t_i^{(2)} &= C_1(t_i^{(1)})
\end{align*}
Where the dimensions of $t_i^{(0)}$ and $t_i^{(1)}$ are in $\mathbb{R}^{4d}$ and that of $t_i^{(2)}$ is in $\mathbb{R}^{d}$. $\mathrm{ReLU}$ is the rectified linear activation function \citep{glorot2011deep}.

\subsection {Bi-affine Pairwise Scores}

We project each contextually encoded token $b_i^{(B)}$ through two separate MLPs to generate two new versions of each token corresponding to whether it will serve as the first or second argument of a relation.
\begin{align*}
e_{i_{head}}^{(0)} &= \mathrm{ReLU}(W_{head}^{(0)} b_i^{(B)}) \\
e_{i_{head}}^{(1)} &= W_{head}^{(1)}e_{i_{head}}^{(0)}
\end{align*}

For each head, tail, relation triple, we calculate a score using a bi-affine operator to create an $n \times n \times l$ tensor $A$ of pairwise affinity scores:
\begin{align*}
a_{ij}=(E_{head}L)e_{tail} + (E_{head}l_b) 
\end{align*}
where $L$ is a $d \times d \times l$ tensor, a learned embedding matrix for each of the $l$ relations.

\subsection {Entity Level Prediction \label{sec:entity_pred}}
Our data is weakly labeled in that there are labels at the entity level but not the mention level, making the problem a form of strong-distant supervision \citep{distant_supervision}. In distant supervision, edges in a knowledge graph are heuristically applied to sentences in an auxiliary unstructured text corpus --- often applying the edge label to all sentences containing the subject and object of the relation. Because this process is imprecise and introduces noise into the training data, methods like multi-instance learning were introduced \citep{riedel2010modeling, surdeanu2012multi}. In multi-instance learning, rather than looking at each distantly labeled mention pair in isolation, the model is trained over the aggregate of these mentions and a single update is made. More recently, the weighting function of the instances has been expressed as neural network attention \citep{verga-mccallum:2016:W16-13, lin2016neural, yaghoobzadeh-adel-schutze:2017:EACLlong}. 

We aggregate over all $M$ representations for each mention pair in order to produce per-relation scores for each entity pair. For each entity pair $ep_i = \{e_{{head}_j},e_{{tail}_k}\}$, we select the $M$ vectors $m_i$ in $A$ where $e_{head}=e_{{head}_j}$, $e_{{tail}}=e_{{tail}_k}$:
\begin{align*}
score_{ep_i} = \log(\sum_i(\exp(m_i)))
\end{align*}

The LogSumExp scoring function is a smooth approximation to the max function and has the benefits of aggregating information from multiple predictions and propagating dense gradients as opposed to the sparse gradient updates of the max \citep{das-EtAl:2017:EACLlong1}.

\subsection{Named Entity Recognition}
\label{sec:model_NER}
In addition to making pair level relation predictions, the final transformer output $b_i^{(B)}$ can be used to make entity type predictions. Our model uses a linear classifier which takes $b_i^{(B)}$ as input and predicts the entity label for each token to produce per-class scores $c_i$:

\begin{align*}
c_i = W_{N} b_i^{(B)}
\end{align*}

We encode entity labels using the BIO encoding. We apply tags to the byte-pair tokenization by treating each sub-word within a mention span as an additional token with a corresponding B or I label.

We train the NER and relation objectives jointly, sharing all embeddings and Transformer parameters. We penalize the named entity updates with a hyperparameter $\lambda$.






\section{Experiments}
\label{sec:results}

We perform experiments on the Biocreative V chemical disease relation extraction (CDR)\footnote{\url{http://www.biocreative.org/}} dataset \citep{li2016biocreative,wei2016assessing}. The dataset was derived from the Comparative Toxicogenomics Database (CTD) which curates interactions between genes, chemicals, and diseases \citep{davis2008comparative}.  These annotations are only at the document level and do not contain mention annotations. The CDR dataset is a subset of these original annotations supplemented with human annotated, entity linked mention annotations. The relation annotations in this dataset are also at the document level only. In addition to the gold CDR data, \citet{peng2016improving} add 15,448 additional PubMed abstracts annotated in the CTD dataset. We consider this same set of abstracts as additional training data (which we denote +Data). Since this data does not contain entity annotations, we take the annotations from Pubtator \citep{pubtator2013}, a state of the art biological named entity tagger and entity linker. In our experiments we only evaluate our relation extraction performance and all models (including baselines) use gold entity annotations for predictions. We compare against the previous best reported results on this dataset not using knowledge base features.~\footnote{The highest reported score is from \cite{peng2016improving} but uses explicit lookups into the CTD knowledge base for the existence of the test entity pair.} Each of the baselines are ensemble methods that make use of additional parse and part-of-speech features. \citet{gu2017chemical} use a CNN sentence classifier while \citet{zhou2016exploiting} use an LSTM. Both make cross-sentence predictions with featurized classifiers.

\subsection{Results}
In Table \ref{results:CDR} we show results outperforming the baselines despite using no syntactic or linguistic features. We show performance averaged over 20 runs with 20 random seeds as well as an ensemble of their averaged predictions. We see a further boost in performance by adding in the additional weakly labeled data. Table \ref{results:cdr_ablation} shows the effects of removing pieces of our model. `CNN only' removes the multi-head attention component from the transformer block, `no width-5' replaces the width-5 convolution of the feed-forward component of the transformer with a width-1 convolution and `no NER' removes the named entity recognition multi-task objective (section \ref{sec:model_NER}). 

\begin{table}
\parbox{.45\linewidth}{
\centering
\begin{tabular}{llll}
    Model & P & R & F1 \\ \hline \hline
    \citet{gu2016chemical}      &  62.0  & 55.1   & 58.3 \\
    \citet{zhou2016exploiting}  &  55.6  & 68.4   & 61.3  \\
    \citet{gu2017chemical}      &  55.7  & 68.1   & 61.3 \\
        \hline
    BRAN                        &  55.6 & 70.8 &  \textbf{62.1} $\pm$ 0.8  \\
    + Data                      &  64.0  & 69.2 & \textbf{66.2} $\pm$ 0.8 \\
    \hline
    BRAN(ensemble)              &  63.3 & 67.1 &  {65.1}  \\
    + Data                      &  65.4 & 71.8 &  \textbf{68.4}  \\
    \hline 
  \end{tabular}
  \caption{Precision, recall, and F1 results on the Biocreative V CDR Dataset.\label{results:CDR}}
  }
\hfill
\parbox{.45\linewidth}{
\centering
\begin{tabular}{llll}
    Model & P & R & F1 \\ \hline \hline
    BRAN (Full)                 &  55.6 & 70.8 &  \textbf {62.1} $\pm$ 0.8  \\
    -- CNN only                 &  43.9  & 65.5 & 52.4 $\pm$ 1.3 \\
    -- no width-5               &  48.2  & 67.2 & 55.7 $\pm$ 0.9  \\
    -- no NER                   &  49.9  & 63.8 & 55.5 $\pm$ 1.8 \\
        \hline 
  \end{tabular}
  \caption{Results on the Biocreative V CDR Dataset showing precision, recall, and F1 for various model ablations. \label{results:cdr_ablation}}
}
\vspace{-.75cm}
\end{table}

\subsection{Implementation Details}
\label{sec:data_processing}

The CDR dataset is concerned with extracting only chemically induced disease relationships (drug-related side effects and adverse reactions) concerning the most specific entity in the document. For example `tobacco causes cancer' could be marked as false if the document contained the more specific `lung cancer.' This can cause true relations to be labeled as false, harming evaluation performance. To address this we follow \cite{gu2016chemical, gu2017chemical} and filter hypernyms according to the hierarchy in the MESH controlled vocabulary~\footnote{\url{https://www.nlm.nih.gov/mesh/download/2017MeshTree.txt}}. All entity pairs within the same abstract that do not have an annotated relation are assigned the NULL label.

The model is implemented in Tensorflow \citep{tensorflow2015-whitepaper}. The byte pair vocabulary is generated over the training dataset -either just the gold CDR data with budget 2500 or gold CDR data plus additional data from section \citep{peng2016improving} with budget 10000. All embeddings are 64 dimensional. Token embeddings are pre-trained using skipgram \cite{mikolov2013efficient} over a random subset of 10\% of all PubMed abstracts with window size 10 and 20 negative samples. The number of transformer block repeats is $B=2$ . We optimize the model using Adam \cite{kingma2014adam} with best parameters chosen for $\epsilon$, $\beta_1$, $\beta_2$ chosen from the development set. The learning rate is set to $0.0005$ and batch size 32. In all of our experiments we set the number of attention heads to $h=4$. 

We clip the gradients to norm 10 and apply noise to the gradients \cite{neelakantan2015adding} with $\eta=.1$. We tune the decision threshold and perform early stopping on the development set. We apply dropout \cite{srivastava2014dropout} to the input layer randomly replacing words with a special UNK token with keep probability $.85$. We additionally apply dropout to the input $T$ (word embedding + position embedding), interior layers, and final state. At each step, we randomly sample a positive or negative (NULL class) minibatch with probability $0.5$.  We merge the train and development sets and randomly take 850 abstracts for training and 150 for early stopping. Our reported results are averaged over 10 runs and using different splits. All baselines train on both the train and development set.

\section{Related work}
Relation extraction is a heavily studied area in the NLP community. Most work focuses on news and web data \citep{ACE2004,riedel2010modeling,hendrickx2009semeval}\footnote{\url{https://tac.nist.gov}}.  There is also a considerable body of work in supervised biological relation extraction including protein-protein \citep{pyysalo2007bioinfer,poon2014distant, mallory2015large}, drug-drug \citep{segurabedmar-martinez-herrerozazo:2013:SemEval-2013}, and chemical-disease \citep{gurulingappa2012development,li2016biocreative} interactions, and more complex events \citep{kim2008corpus, riedel-EtAl:2011:BioNLP-ST}. Recent neural network approaches to relation extraction have focused on CNNs \citep{dossantos-xiang-zhou:2015:ACL-IJCNLP,zeng-EtAl:2015:EMNLP} or LSTMs \citep{miwa-bansal:2016:P16-1, verga-EtAl:2016:N16-1,zhou-EtAl:2016:P16-2} and replacing stage-wise information extraction pipelines with a single end-to-end model \citep{miwa-bansal:2016:P16-1, Ammar2017ai2semeval,li2017neural}.

A few exceptions exist that perform cross-sentence relation extraction \citep{swampillai-stevenson:2011:RANLP,quirk-poon:2017:EACLlong, peng2017cross}. Most similar to our work would be \citep{peng2017cross} which uses a variant of an LSTM to encode document-level syntactic parse trees. Our work differs in several key ways. It operates over raw tokens negating the need part of speech or parse features which can lead to cascading errors. We also use a feed-forward neural architecture which encodes long sequences far more efficiently compared to the graph LSTM network of \citep{peng2017cross}. Finally, our model considers all mention pairs rather than a single mention pair at a time.

Pairwise bilinear models have also been used extensively in knowledge graph link prediction \citep{rescal,li-EtAl:2016:P16-14} sometimes restricting the bilinear relation matrix to be diagonal \citep{distmult} or diagonal and complex \citep{trouillon2016complex}. Our model is similar to recent approaches in neural graph-based parsing where bilinear parameters are use to score a head-dependent relationship \citep{kipperwasser-goldberg-parser, dozat2016deep}.

\section{Conclusion}

We present a bilinear relation attention network that simultaneously produces predictions for all mention pairs within a document. With this model we are able to outperform the previous state of the art on the Biocreative V CDR dataset. Our model also lends itself to other tasks such as hypernym prediction, coreference resolution, and entity resolution. We plan to investigate these directions in future work.

\bibliography{sources}
\bibliographystyle{emnlp_natbib}

\end{document}